\documentclass[manuscript,nonacm,hidelinks]{acmart}

\usepackage{amsmath}
\usepackage{booktabs}
\usepackage{algorithm}
\usepackage{algpseudocode}
\usepackage{graphicx}
\usepackage{listings}
\usepackage{xcolor}
\usepackage{tabularx}
\usepackage{array}
\usepackage{siunitx}
\usepackage{tikz}
\usepackage{pgfplots}
\pgfplotsset{compat=1.18}
\usepackage{subcaption}
\usepackage[utf8]{inputenc}

\renewcommand\footnotetextcopyrightpermission[1]{}
\author{Pratik Poudel}
\affiliation{%
  \institution{Florida International University}
  \city{Miami}
  \state{Florida}
  \country{USA}
}
\email{ppoud001@fiu.edu}

\title{Stateful KV Cache Management for LLMs: Balancing Space, Time, Accuracy, and Positional Fidelity}

\title{Stateful KV Cache Management for LLMs: Balancing Space, Time, Accuracy, and Positional Fidelity}

\begin{document}

\begin{abstract}
The Key-Value (KV) cache is integral to efficient autoregressive inference in Large Language Models (LLMs), yet its unbounded growth in stateful multi-turn scenarios presents significant challenges. This paper investigates the critical interplay between KV cache management strategies, the inherent architectural context limits of models like `meta-llama/Meta-Llama-3-8b-instruct`, and the often-overlooked integrity of positional encodings. Through empirical analysis using a stateful benchmarking framework on extended conversational data, we demonstrate that LLM generation quality severely degrades when the accumulated KV cache approaches or exceeds the model's pre-trained architectural context window (e.g., 8192 tokens for Llama 3), a failure mode distinct from mere GPU memory exhaustion. Our findings reveal that common eviction strategies, even those designed to retain a high percentage of tokens (e.g., 99\% via AttentionTop), can paradoxically worsen performance if they disrupt the positional coherence of the cached states, especially when applied to already long contexts. This disruption occurs because LLMs rely on consistent positional signals (e.g., RoPE) learned from continuous text; compacting a cache by removing non-contiguous tokens scrambles these signals, leading to model confusion and degenerative output. We further analyze intra-turn cache dynamics, highlighting how prefill phases can significantly inflate cache size before generation, complicating eviction effectiveness. Conversely, we show that simpler strategies preserving contiguous blocks of context (e.g., retaining only an initial "gist") can maintain superior coherence compared to strategies that cause positional disruption or a baseline that allows the cache to exceed architectural limits. This work underscores the necessity for eviction techniques that are acutely aware of model-specific context limits, the nuances of positional encoding, and the full cycle of cache changes within conversational turns, advocating for a more holistic view of "cache health" beyond mere size.
\end{abstract}


\maketitle
\section{Introduction}
Large Language Models (LLMs) have demonstrably revolutionized natural language processing, powering increasingly sophisticated applications from advanced conversational agents and AI co-pilots to tools for long-form content generation and analysis \cite{vaswani17attention, brown20gpt3}. A cornerstone of their efficient autoregressive inference is the Key-Value (KV) cache, which stores past attention states to prevent costly re-computation. However, in stateful, multi-turn interactions—the bedrock of engaging AI experiences—this cache accumulates information with each turn, growing linearly with the conversational context. This unchecked growth presents a dual crisis: firstly, the tangible risk of exhausting finite GPU memory, and secondly, a more insidious threat – the degradation of the model's generation quality as the effective sequence length in its cache approaches or, critically, exceeds its pre-trained architectural context window.

Navigating this challenge requires robust KV cache management. Yet, prevalent eviction strategies often prioritize *what* tokens to retain (e.g., based on recency or attention scores) with less explicit consideration for *how* the resulting compacted cache state interacts with the model's fundamental sequence processing capabilities, particularly its reliance on consistent positional information. This paper investigates these complex interdependencies, moving beyond the traditional Space-Time-Accuracy trilemma to incorporate a fourth crucial dimension: **Positional Fidelity**. Our central thesis, substantiated by empirical analysis of `meta-llama/Meta-Llama-3-8b-instruct` using extended, multi-turn conversational benchmarks, is that **preserving the structural integrity of positional information within the KV cache during eviction is as critical, if not more so, than merely retaining a high percentage of purportedly "important" tokens, especially when the model operates near its architectural context limits.** The expectation that sophisticated, high-retention eviction strategies should inherently preserve performance often belies the reality of how LLMs interpret sequences.

This research uncovers several pivotal insights into stateful KV cache management:
\begin{itemize}
    \item \textbf{Architectural Limits as Hard Boundaries:} We demonstrate that LLMs like Llama 3 exhibit significant, often catastrophic, quality degradation when their stateful KV cache grows beyond their trained architectural context window (e.g., 8192 tokens). This performance collapse occurs irrespective of sheer GPU memory availability, highlighting a fundamental processing constraint tied to how models comprehend sequence order and relationships at scale.
    \item \textbf{The Unseen Burden of Prefill Dynamics:} Our analysis reveals that the common prefill phase—where user input is processed and added to the cache at the start of a turn—can drastically inflate the KV cache size \textit{before} the model even begins generating its response. This often pushes the cache well beyond intended operational thresholds, creating a more challenging environment for subsequent eviction logic and potentially stressing the model from the outset of its generation task.
    \item \textbf{The Paradox of High-Retention Eviction and Positional Scrambling:} We find that common eviction strategies, such as "AttentionTop" (even when configured for very high token retention, e.g., 99\%), can paradoxically induce severe performance drops. This occurs when these strategies, in an attempt to preserve salient tokens from a long context, select and compact non-contiguous token states. Such an operation can effectively "scramble" the positional encodings (e.g., RoPE) that are vital for the model's understanding of sequence order, causality, and narrative flow, leading to confusion and degenerative output.
    \item \textbf{The Efficacy of Positional Coherence:} Conversely, we show that simpler strategies maintaining the contiguity of retained token blocks (e.g., preserving only an initial "gist" of a long conversation) can yield strikingly more coherent and task-relevant outputs. This occurs even when compared against a baseline model struggling with an over-limit context, or against a nearly-full cache that has been positionally compromised by a more complex eviction strategy.
\end{itemize}
Our work, therefore, contributes a deeper understanding of the nuanced failure modes associated with KV cache management in long-context stateful LLMs. It underscores the imperative for developing cache eviction techniques that are acutely cognizant not only of memory constraints and content salience but also—and critically—of the model's architectural context capacity, the underlying mechanics of its positional encoding scheme, and the full cycle of cache modifications within each conversational turn. By illuminating these interactions, we aim to guide the development of more robust strategies that enable truly extended, coherent, and reliable multi-turn dialogues with LLMs.

\section{Background} 
\label{sec:kv-cache}

\subsection{KV Caching in Autoregressive LLMs}
Transformer-based LLMs \cite{vaswani17attention} rely on attention mechanisms that compute pairwise interactions between tokens in a sequence. During autoregressive generation, a naive implementation would recompute all attention patterns for the entire sequence at each step, resulting in quadratic complexity as sequence length increases.

The Key-Value (KV) cache optimization addresses this inefficiency by storing the precomputed key and value projections from previous tokens. For layer $\ell\!\in\!\{1,\dots,L\}$ and head $h\!\in\!\{1,\dots,H\}$, we denote the cached keys and values for the first $t^{\star}$ tokens as $K^{(\ell,h)}_{1:t^{\star}}\!\in\!\mathbb{R}^{t^{\star}\!\times d_{k}}$ and $V^{(\ell,h)}_{1:t^{\star}}\!\in\!\mathbb{R}^{t^{\star}\!\times d_{v}}$, where $d_k$ and $d_v$ are the key and value dimensions. At each new step, only the query for the current token $q^{(\ell,h)}_{t^{\star}+1}\!\in\!\mathbb{R}^{d_{k}}$ needs to be computed, which then interacts with the cached keys and values:

\begin{equation}
\label{eq:attn}
\text{Attn}\!\bigl(q^{(\ell,h)}_{t^{\star}+1},K^{(\ell,h)}_{1:t^{\star}},
                   V^{(\ell,h)}_{1:t^{\star}}\bigr)
=\operatorname{softmax}\!\Bigl(
 \tfrac{q^{(\ell,h)}_{t^{\star}+1} (K^{(\ell,h)}_{1:t^{\star}})^{\!\top}}
      {\sqrt{d_{k}}}\Bigr)V^{(\ell,h)}_{1:t^{\star}} .
\end{equation}

This optimization reduces per-token complexity from $O(t^2)$ to $O(t)$ by avoiding redundant computations. However, it introduces a memory overhead that grows linearly with sequence length. For a model with $L$ layers and $H$ heads, the memory footprint $M_{\text{KV}}$ is:

\begin{equation}
\label{eq:footprint}
M_{\text{KV}} = 2\,L\,H\,T\,d_{k}
\quad\text{elements}\quad
\Rightarrow\quad
B_{\text{KV}} = 2LHTd_{k}b\;\text{bytes},
\end{equation}

where $b$ is the number of bytes per element (typically 2 bytes for half-precision).

As an example, for Llama 2 7B with 32 layers, 32 attention heads, and a head dimension of 128, a sequence of 2048 tokens requires approximately 1 GiB of memory just for the KV cache. This memory footprint increases linearly with sequence length, creating a bottleneck for long-context scenarios.

\subsection{The Context Window Limit and Positional Encodings}
Large Language Models (LLMs) are inherently constrained by a pre-defined maximum context window, such as the 8192 tokens for Llama 3. This limitation is not arbitrary but is often a direct consequence of the chosen positional encoding scheme implemented during their pre-training. Positional encodings are a crucial component of the Transformer architecture, as the self-attention mechanism, by its nature, is permutation invariant – it does not inherently possess information about the order of tokens in a sequence. Without positional information, the model would treat input text as an unordered bag of words, severely limiting its ability to understand grammar, syntax, and semantic relationships that depend on sequence order.

One prevalent and effective scheme is Rotary Positional Embeddings (RoPE) \cite{su2021roformer}. Unlike earlier methods that added sinusoidal functions to the input embeddings or learned positional embeddings, RoPE introduces positional information by rotating pairs of features in the query and key embeddings at each layer of the Transformer. Specifically, for a token at position $m$ in a sequence, its embedding $x_m$
is transformed by a rotation matrix that depends on $m$. This is achieved by element-wise multiplication with complex numbers $e^{im\theta_d}$, where $\theta_d$ are predefined constants.
This formulation ingeniously injects absolute positional information (the specific position m) in a way that promotes relative positional awareness. The dot product attention between a query at position m and a key at position n will then depend on their relative distance m\-n, due to the properties of these rotations. This allows the model to inherently understand the relative distances and order between tokens, which is vital for coherent text generation and comprehension.

During the training phase, the model learns to utilize these injected positional signals to interpret the relationships between tokens based on their positions. The specific maximum context window length is determined at this stage, and the positional encodings are optimized for sequences up to this length.

However, the efficacy of RoPE, like many positional encoding schemes, diminishes when the actual sequence length S processed by the model, particularly within its Key-Value (KV) cache, surpasses this pre-trained maximum. The KV cache stores the key and value vectors from the self-attention layers for previously processed tokens, enabling efficient generation of subsequent tokens. When S exceeds the trained limit, the rotational angles in RoPE may extrapolate beyond the range they were optimized for. This extrapolation can lead to unpredictable and often degraded representations of positional information. The model's learned ability to interpret token order and relative distances becomes compromised, resulting in a breakdown in its capacity to process the extended context coherently. As will be discussed in a later section, this phenomenon significantly impacts the inference process, particularly how the KV cache, upon reaching and exceeding the context window, can corrupt future response generation due to this breakdown in positional understanding.

\subsection{The Challenge of Stateful KV Cache Eviction in Extended Conversation}
In stateful, multi-turn conversational AI systems, the Key-Value (KV) cache S grows with each new exchange. To avoid exceeding architectural limits or depleting GPU memory, we must employ effective eviction strategies that selectively remove token states from the cache. The success of these strategies depends on overcoming several key challenges. The core function of these eviction strategies is to judiciously select and discard token states from the KV cache. However, the efficacy of such strategies is contingent upon the successful mitigation of several critical challenges.

The primary challenge resides in minimizing \textit{Information Loss}. This refers to the inadvertent discarding of token states that possess significant referential or contextual importance for the model's comprehension and generation of coherent responses in subsequent conversational turns. The premature eviction of such critical tokens can lead to a degradation in the quality, relevance, and factual accuracy of the model's output.

A second, and equally significant, challenge is the mitigation of \textit{Positional Disruption}. The cached token states inherently retain positional information, often encoded through positional encodings, which informs the model about the sequential relationships between tokens. Certain eviction methodologies, particularly those that do not preserve the contiguity or relative ordering of the cached states, risk rendering these original positional encodings misleading. This discrepancy can confuse the model's attention mechanisms, potentially leading to incoherent or contextually inappropriate outputs, especially as the conversational history, and thus the KV cache, becomes extensive.

Finally, the \textit{Computational Overhead} associated with the eviction process itself presents a practical challenge. For real-time conversational applications, the algorithms governing token eviction must be computationally lean and execute with minimal latency. Complex eviction strategies, while potentially more precise in identifying less critical tokens, may introduce unacceptable delays, thereby compromising the user experience.

\section{Related Work}

The challenges of deploying Large Language Models (LLMs) for long-context, stateful inference, particularly concerning the KV cache, have spurred considerable research. Efforts can be broadly categorized into: (1) foundational architectural and encoding enhancements to improve inherent model capabilities for longer sequences, (2) techniques to reduce the static memory footprint of the KV cache, and (3) dynamic strategies for managing the KV cache contents during inference. Our work focuses primarily on the third category, with a specific emphasis on the under-explored impact of positional disruption in stateful eviction.

\subsection{Enhancing Model Capabilities for Extended Contexts}

This section addresses foundational improvements to model architectures and positional encodings that enhance the inherent capability of LLMs to process longer sequences.

\subsubsection{Architectural Innovations for Sequence Processing}

Early efforts to handle longer sequences involved fundamental changes to the Transformer architecture aimed at reducing the quadratic complexity of self-attention. Sparse Attention mechanisms \cite{child2019generating, beltagy2020longformer} restrict attention to local windows or predefined patterns, while Linear Attention variants \cite{katharopoulos2020transformers, wang2020linformer} reformulate attention to achieve linear complexity. While these modifications can reduce computational load and enable processing of longer sequences, they may also alter the information flow and representation capabilities of the model.

\subsubsection{Advancements in Positional Encodings and Context Window Extension}

The pre-defined context window of LLMs is often tied to the limitations of their positional encoding schemes, such as Rotary Positional Embeddings (RoPE) \cite{su2021roformer}. Significant research has focused on extending these limits. Techniques like Positional Interpolation (PI) \cite{chen2023extending} and YaRN (Yet another RoPE extensioN method) \cite{peng2023yarn} adapt RoPE to accommodate longer sequences than seen during pre-training by modifying interpolation or scaling strategies. Alternative positional encoding schemes, such as ALiBi (Attention with Linear Biases) \cite{press2021train}, have been proposed for their superior extrapolation properties, allowing models to generalize to longer contexts without explicit fine-tuning for such lengths. Other approaches include fine-tuning models on progressively longer sequences \cite{xiong2024effective} or employing specialized training objectives to improve long-context understanding.

While these methods expand the theoretical context capacity, the practical memory footprint of the KV cache remains a bottleneck, necessitating direct memory management techniques.

\subsection{Reducing the KV Cache Memory Footprint}

This section covers techniques aimed at reducing the memory footprint of the KV cache, thereby allowing for longer sequences to be processed within given hardware constraints.

\subsubsection{Cache-Focused Architectural Modifications}

To directly reduce the size of the KV cache, architectural changes have been proposed. Multi-Query Attention (MQA) \cite{shazeer2019fast} significantly reduces memory by sharing key and value projections across all attention heads. Grouped-Query Attention (GQA) \cite{ainslie2023gqa} offers a balance by grouping heads that share projections, allowing a trade-off between model quality and memory efficiency, as seen in models like Llama 2 \cite{touvron2023llama}.

\subsubsection{KV Cache Quantization}

Another direct approach to reduce the memory footprint is by decreasing the precision of the stored keys and values. Post-Training Quantization methods specifically targeting the KV cache, such as KV-Quant \cite{hooper2023kv}, apply techniques to represent cached states using fewer bits (e.g., 8-bit, 4-bit). Outlier-aware quantization attempts to preserve model performance by maintaining higher precision for statistically significant values, while hardware-aware quantization optimizes schemes for specific hardware.

\subsection{Dynamic Management of the KV Cache via Eviction and Prioritization}

When the KV cache approaches its capacity limits, particularly in stateful, multi-turn conversations, dynamic strategies are required to decide which token states to retain or discard. These approaches aim to balance information preservation against memory constraints.

\subsubsection{Token Pruning and Eviction Strategies}

Various heuristics and learned methods have been proposed for token eviction. Window-Based Methods retain a sliding window of the most recent tokens, sometimes with modifications like 'attention sinks' to preserve crucial initial tokens \cite{zhang2023h2o, xiao2023streamingllm}. Attention-Based Methods leverage the model's own attention scores or learned importance metrics \cite{khandelwal2023prompt, zhang2024retentive}, aiming for a more informed selection of tokens to prune. Compression-Based Approaches compress multiple token states into more compact representations. Hybrid Approaches combine multiple criteria, such as recency, attention, and positional heuristics \cite{ren2023ds, zhang2023efficiently}.

However, as noted in Section 2.3, while these methods address information loss and computational overhead, the explicit analysis of their impact on positional disruption-and consequently, the integrity of positional encodings like RoPE post-eviction-particularly in stateful, long-context conversational settings, remains less developed. This is a central focus of our work.

\subsubsection{Hierarchical Storage Solutions}

To virtually expand the KV cache beyond fast GPU memory, hierarchical storage techniques offload less critical or older entries to slower, larger memory tiers like CPU RAM or disk \cite{kwon2023efficient, pope2023efficiently, zhang2023h2o}. These methods often involve sophisticated caching policies and prefetching but still rely on effective identification of which tokens can be moved to slower tiers without significantly impacting performance.

\section{Methodology: Stateful Benchmarking of Eviction Strategies}
\label{sec:methodology}

\subsection{Experimental Setup and Model}
Our benchmark investigates stateful inference performance using the \textit{Meta-Llama-3-8b-instruct} model. Experiments were conducted on a server equipped with NVIDIA A100 80GB PCIe GPUs. In this stateful paradigm, the KV cache, represented by the \textit{past\_key\_values} object, is maintained and passed between consecutive turns of a single conversational item. For our tests, we utilized a subset of the ShareGPT dataset, specifically selecting and adapting conversational items to ensure extended dialogues, often exceeding 30+ turns, which are designed to definitively surpass the Llama 3 8B model's 8192-token architectural context window. The KV cache is only reset upon the commencement of an entirely new conversational item. This methodology allows for the precise observation of cumulative KV cache growth and the impact of various eviction strategies on this evolving state throughout a long conversation.
The primary metrics collected per turn encompass:
\begin{itemize}
\item \textbf{KV Cache Size (MB):} Measured directly from GPU memory. This includes measurements after processing the current turn's input (prefill phase) and after each token is generated during the decoding phase, reflecting the memory consumed by Hugging Face's DynamicCache object.
\item \textbf{Latency Metrics:} This includes Time To First Token (TTFT) for the current turn and the generation throughput in tokens per second during the decoding phase.
\item \textbf{Generation Quality:} Assessed using a dual approach: quantitatively by an LLM-as-a-judge (GPT-4o via Azure OpenAI API) scoring responses on a 1-10 scale for coherence, relevance, and helpfulness, and qualitatively through manual analysis of the generated text for task completion and logical consistency.
\item \textbf{Eviction Statistics:} Number of tokens evicted and the computational time taken for the eviction process, where applicable.
\item \textbf{Overall Throughput:} Total number of tokens generated per second for the entire turn's generation phase.
\end{itemize}
A \textit{kv\_threshold\_mb} was established (e.g., 600MB, which corresponds to approximately 5600 tokens for the Llama 3 8B model) to trigger the eviction strategies. Eviction occurs if the cache size, after processing the current turn's input tokens or during the generation of new tokens, exceeds this predefined limit.

\subsection{Evaluated KV Cache Management Strategies}
We implemented and systematically evaluated the following KV cache management strategies within our stateful benchmarking framework:
\begin{itemize}
\item \textbf{Baseline (No Eviction):} This strategy does not apply any eviction. The KV cache is allowed to grow unbounded until it either reaches the GPU's memory capacity or, more critically for our study, far exceeds the model's architectural context window. This serves as a crucial control to observe the model's behavior under extreme context length pressure and provides the reference against which other strategies are compared, particularly concerning performance degradation once architectural limits are breached.rem
\item \textbf{AttentionTop (Attention-Based Eviction):} This strategy aims to retain a \textit{keep\_ratio} of tokens that exhibit the highest cumulative attention scores from the most recent model pass. The attention scores are carefully processed to account for both the prefill phase (where multiple input tokens are processed simultaneously) and the decoding phase (where tokens are generated one by one). The goal is to preferentially keep tokens deemed more "important" by the model's own attention mechanism.
\item \textbf{SlidingWindowGist (Hybrid Windowing):} This approach combines recency and primacy. It retains a fixed number of initial tokens (\textit{gist\_token\_count}) from the beginning of the conversation (the "gist") and a fixed number of recent tokens (\textit{recent\_token\_count}) from the end of the current context. Tokens in the middle are discarded. This strategy tests the hypothesis that the initial context and the most recent exchanges are often most critical for maintaining conversational flow.
\item \textbf{EvictOldest (Sliding Window / FIFO):} Not explicitly detailed in the provided text but present in the line plot and important for context. This strategy typically maintains a fixed-size window of the most recent tokens by evicting the oldest tokens once the window capacity or memory threshold is exceeded. It's a common approach prioritizing recency.
\end{itemize}
All strategies that involve modification of the cache operate by creating new lists of key and value tensors containing only the selected token states. These new tensors are then used to update the existing \textit{DynamicCache} object. This ensures that internal properties of the cache, such as \textit{seen\_tokens} (the count of tokens currently in the cache), accurately reflect its new, post-eviction length and state, which is crucial for the correct functioning of positional encodings in subsequent generation steps.

\section{Results and Analysis: The Importance of Positional Encodings}
\label{sec:results}

\begin{figure}[h]
\centering
\includegraphics[width=0.9\linewidth]{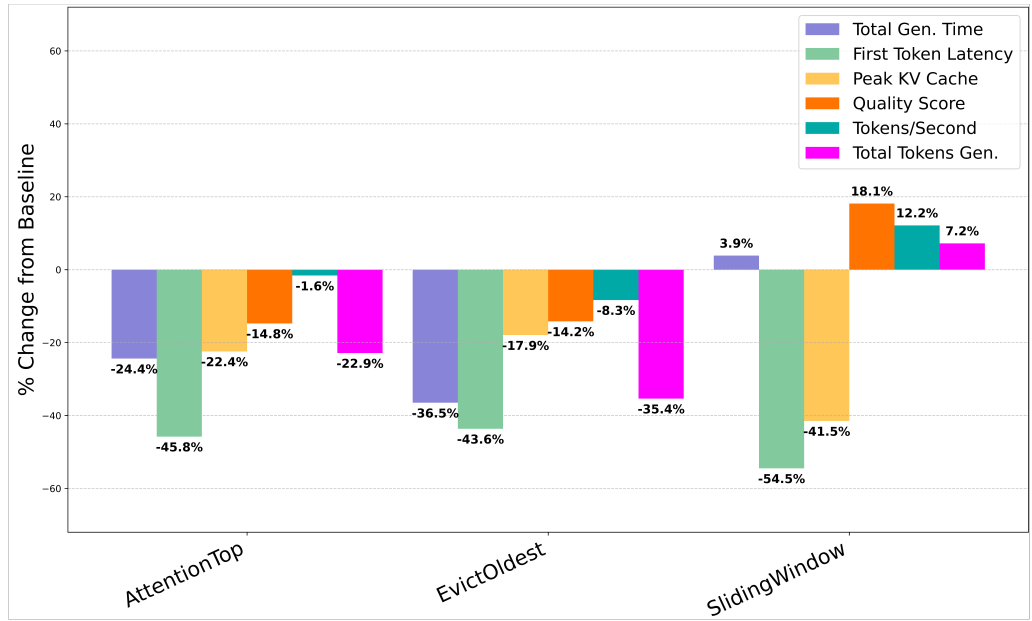} 
\caption{Performance Metrics Comparison (\% Change from Baseline) with KV cache management strategies. Negative percentage means lower values for the metric compared to baseline.\label{fig:perfomance_comparision}}
\vspace{-1em}
\end{figure}

Our experiments with \textit{Meta-Llama-3-8b-instruct} in a stateful multi-turn setting, particularly using extended dialogues from the ShareGPT dataset designed to push beyond its 8192-token context window, yielded critical insights into the behavior of LLMs with long contexts and the impact of KV cache eviction.

\subsection{Baseline Performance and Architectural Limits}
The Baseline strategy, configured with a very high \textit{kv\_threshold\_mb} (e.g., 10000MB, effectively disabling programmatic eviction within typical conversational lengths for this study), initially demonstrated coherent generation. However, as the accumulated KV cache inexorably grew with each turn, approaching and then significantly exceeding the model's 8192-token architectural limit (which corresponds to roughly 1024MB of KV cache for Llama 3 8B), a pronounced degradation in output quality was consistently observed. For instance, in one representative multi-turn dialogue, by Turn 9/10, where the accumulated cache contained over 8800 token states, the model began to exhibit failure modes such as repeating the user's prompt verbatim. By Turn 20, with the cache exceeding 15000 tokens, responses typically degenerated into incoherent, repetitive character sequences or nonsensical phrases. This confirms that exceeding the architectural context length induces model failure due to an inability to process such long sequences coherently, irrespective of the sheer availability of GPU memory to hold the oversized cache.

\subsection{KV Cache Dynamics and Eviction Threshold Interaction}
\label{sec:kv_dynamics_detailed}

\begin{figure}[htb]
\centering
\includegraphics[width=\linewidth]{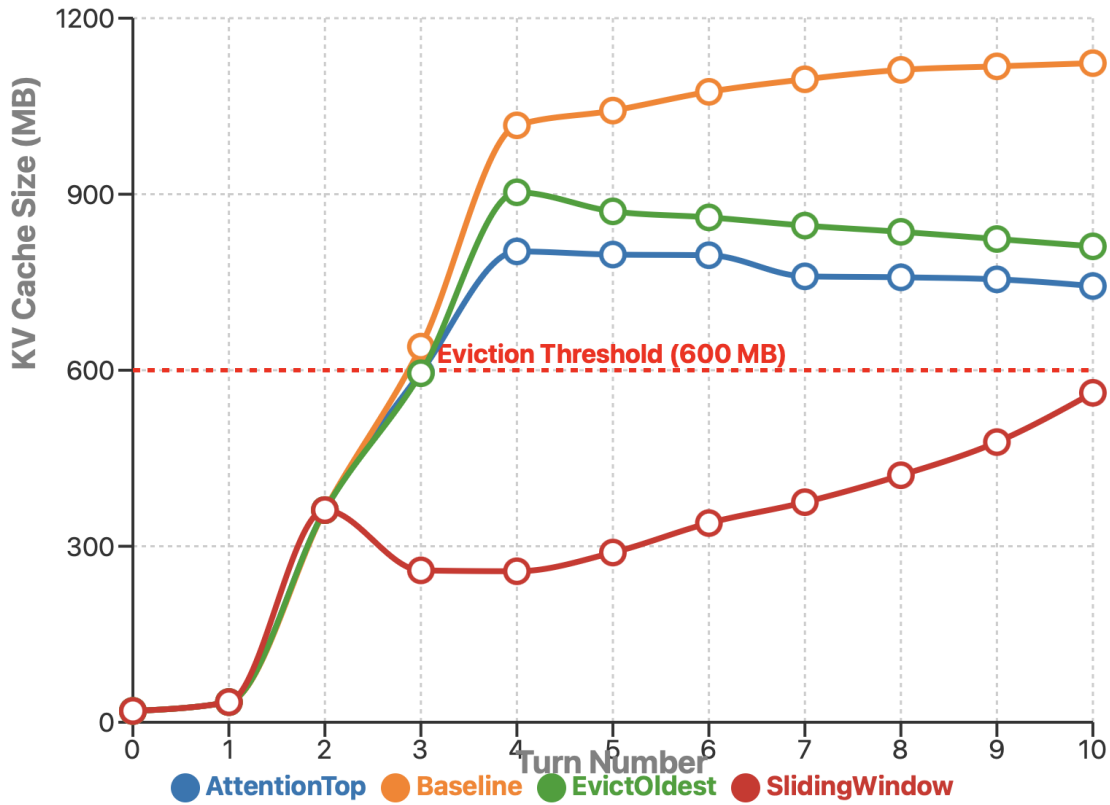} %
\caption{KV Cache Growth per Turn for a conversation with 50+ turn reported only till turn 10(During end of Generation Phase). The cache eviction threshold is 600MB . Note how AttentionTop and EvictOldest operate above this threshold after Turn 3, reflecting complex intra-turn dynamics.}
\label{fig:kv_cache_eog_growth}
\end{figure}

Figure \ref{fig:kv_cache_eog_growth} presents the KV cache size (in MB) at the end of the generation phase for each conversational turn, comparing the Baseline, SlidingWindow, EvictOldest, and AttentionTop strategies. The dashed red line indicates a 600MB eviction threshold, which is illustrative of a memory constraint that might trigger eviction (our actual experimental trigger was 600MB).

The \textbf{Baseline} strategy (labeled 'Baseline' in Figure \ref{fig:kv_cache_eog_growth}, typically orange or similarly distinct) exhibits continuous, linear growth, surpassing the 600MB threshold around Turn 3, visually representing the uncontrolled accumulation of past token information.

More revealingly, both the \textbf{AttentionTop} (e.g., blue line) and \textbf{EvictOldest} (e.g., green line) strategies also cross the 600MB threshold around Turn 3 or 4. Notably, they remain significantly above this illustrative threshold for several subsequent turns before showing a gradual decrease after Turn 5 or 6. This behavior, which might seem counter-intuitive for eviction strategies designed to manage cache size, is a consequence of the interplay of several mechanics within each turn:

\begin{itemize}
\item \textbf{Eviction Triggering and Conservative Rates:} In our stateful setup, eviction is triggered \textit{before} the prefill phase of a new turn if the KV cache size from the \textit{end of the previous turn} exceeds the configured threshold (600MB). However, the eviction mechanisms themselves, especially with high retention targets (e.g., AttentionTop aiming to retain 99\% of tokens), or low discard rates per step, mean that the initial reduction in cache size might be modest relative to the overall cache. If the cache is already substantially over the threshold, a small percentage reduction may not bring it significantly below.
\item \textbf{Impact of the Prefill Phase:} This is a critical factor. \textit{After} any pre-turn eviction, the prefill phase processes the current turn's user input. This phase \textit{always} adds all new user tokens to the KV cache. In dialogues with substantial user input, especially in early to mid-turns (e.g., the transition from Turn 3 to Turn 4 depicted by the sharp rise in Figure \ref{fig:kv_cache_eog_growth}), this prefill stage can cause a large surge in the KV cache size. This surge pushes the cache well above the threshold \textit{again}, even before the model begins generating its response for the current turn. The end-of-generation values plotted in Figure \ref{fig:kv_cache_eog_growth} inherently reflect this post-prefill, post-generation state.

\item \textbf{Token Additions During Generation:} Throughout the generation phase of the current turn, new tokens (the model's response) are continuously added to the cache. If the cache size remains above the threshold, the eviction policy is applied concurrently or iteratively. However, if the rate of token addition during generation, coupled with the already high starting point of the cache post-prefill, surpasses the rate of token removal by the eviction strategy (due to high retention targets or other limiting factors in the eviction logic), the cache size will remain elevated or even continue to grow. This explains the sustained high cache levels for AttentionTop and EvictOldest between Turns 4 and 6.

\end{itemize}

The observed decrease in cache size for AttentionTop and EvictOldest after Turn 5/6, as seen in Figure \ref{fig:kv_cache_eog_growth}, signals a shift in these dynamics. This typically occurs when subsequent user inputs become shorter (reducing the prefill surge) and/or the model generates fewer tokens in its response (possibly due to increasing incoherence from operating near its limits, or issues like the "token encoding mismatch" you hypothesized). In such scenarios, the consistent (albeit proportionally small) eviction action begins to outweigh the reduced rate of new token additions, leading to a net decrease in the cache size per turn.

In contrast, the \textbf{SlidingWindow} strategy (red line in Figure \ref{fig:kv_cache_eog_growth}), by strictly maintaining a fixed-size window of the most recent tokens (once the window is full), demonstrates more consistent control over its cache size after an initial growth phase. It tends to fluctuate at a lower cache size compared to AttentionTop and EvictOldest under the depicted conditions, as its eviction is more directly tied to a fixed token count rather than a percentage of a potentially very large and fluctuating base.

These detailed observations underscore that an eviction threshold often functions as a trigger point rather than a hard ceiling, particularly when substantial new data is introduced during prefill phases common in conversational AI. The cache size measured at the end of a generation turn is a cumulative result of these complex intra-turn dynamics. This understanding is paramount when evaluating the efficacy and failure modes of various eviction strategies, especially concerning their ability to not only manage memory but also maintain the necessary conditions for coherent model generation when contexts become very long.

\subsection{Impact of AttentionTop Eviction with High Retention}
\label{sec:attentiontop_impact_revised}
Setting the \textit{kv\_threshold\_mb} = 600MB (approx. 5600 tokens) allowed us to test eviction strategies on contexts that were already very long, potentially stressing the model's coherence, yet before the absolute architectural limit of 8192 tokens was breached by the prefill phase. For the "AttentionTop" strategy, configured with a \textit{keep\_ratio} = 0.99 (intending to retain 99\% of tokens):
\begin{itemize}
\item \textbf{Initial Coherence:} In early turns (e.g., Turns 1-3), where the accumulated cache after adding the current turn's input remained below the 600MB threshold, no eviction occurred, and model responses were generally coherent and task-appropriate.
\item \textbf{Degradation Post-Eviction Trigger:} As discussed in Section \ref{sec:kv_dynamics_detailed}, when the cache size did exceed the threshold (e.g., around Turn 4), eviction was triggered. However, due to the significant cache size increase post-prefill (evident in Figure \ref{fig:kv_cache_eog_growth}), the model was already operating on a very large context (e.g., >800MB or >6400 tokens for AttentionTop at Turn 4's end). Even applying a 99\% keep ratio to such a large base meant that the model still had to process a substantial, and now potentially fragmented, cache. The subsequent generation quality often dropped significantly. For instance, if logs indicated that an effective 10\% keep ratio (a more aggressive eviction) under certain conditions led to gibberish, it underscores the sensitivity. Even a 99\% keep ratio applied to a cache near 5600 tokens (reducing it to ~5320 tokens) sometimes resulted in similar degradation if the context was already stressed or positionally compromised.
\item \textbf{The "Part 3" Prompt Scenario:} A particularly illustrative case involved a long user prompt ("Part 3...") being introduced when the cache was already substantial (around 5500 tokens pre-prompt). The prefill phase for this long prompt pushed the total cache size to near or slightly over 8000 tokens. The model's initial response segment was coherent but exhibited trailing repetition. On the \textit{next} turn, after the AttentionTop strategy (with its 0.99 keep\_ratio) was applied to this ~8000+ token cache (resulting in a cache of ~7600+ tokens), the model's response to a simple subsequent prompt (e.g., "provide an elevator pitch") degenerated into highly repetitive, incoherent output.
\end{itemize}
This pattern demonstrates that merely retaining a high percentage of tokens, even those with high attention scores, is insufficient if the context is already extremely long and approaches the model's architectural limits. Discarding even 1\% of tokens from such a stressed context can be catastrophic if these tokens were vital for maintaining long-range dependencies or, more critically, if the act of removing them (creating gaps and re-compacting) disrupts the positional integrity of the cached states. The "AttentionTop" strategy, while aiming to preserve locally important tokens, does not inherently guarantee the preservation of global sequence structure or the continuity of positional signals crucial for extended coherence.

\subsection{The Surprising Efficacy of Simple Gist Retention}
In stark contrast, the \textit{SlidingWindowGist} strategy, when configured to retain only the first \textit{gist\_token\_count} = 2000 tokens (and \textit{recent\_token\_count = 0}), demonstrated remarkable resilience. This was tested in scenarios where the Baseline strategy (with its KV cache far exceeding 8000 tokens) was failing (e.g., on Turn 10, merely repeating the user's prompt). In these same situations, the \textit{SlidingWindowGist} strategy, by evicting the cache down to only the initial 2000 tokens, produced a significantly more coherent and task-relevant response to the identical prompt.
\begin{itemize}
\item \textbf{Baseline (Turn 10, >8000 tokens in cache):} User: "Write me a pitch I can use on shark tank for this company." Model: (Repeated the prompt verbatim or produced unrelated content).
\item \textbf{SlidingWindowGist (Turn 10, after eviction to first 2000 tokens from the original long cache):} User: "Write me a pitch I can use on shark tank for this company." Model: "Here's a pitch for DaySculpting on Shark Tank..." (Followed by a full, creative, and contextually relevant pitch).
\end{itemize}
This outcome strongly suggests that providing the model with a shorter, but positionally intact and foundationally relevant segment of context (the initial 2000 tokens in this case) is far more effective for maintaining coherence than forcing it to operate on an overly long context that has either exceeded its architectural limits or has been slightly reduced but positionally compromised by non-contiguous eviction. The 2000-token gist, despite omitting a large swath of intermediate conversational history (over 6000 tokens), retained sufficient core information and, crucially, its original positional structure, enabling the model to successfully address the "Shark Tank pitch" prompt in a meaningful way.

\section{Discussion: Positional Integrity as a Key Factor}
\label{sec:discussion_revised}
Our findings consistently underscore that the architectural context limit of an LLM (e.g., 8192 tokens for Llama 3) is a critical performance boundary, not merely a suggestion. As demonstrated, once the effective context within the KV cache approaches or exceeds this limit, generation quality degrades rapidly and often catastrophically, leading to common failure modes such as prompt repetition, vacuous responses, or incoherent output. This degradation occurs even when ample GPU memory is available to store the oversized cache, pointing to a fundamental limitation in the model's ability to process and make sense of information beyond its trained sequence length, rather than a simple memory capacity issue for the cache itself.

The more striking and central observation of our study is the profound impact of eviction strategies on already extended contexts, particularly concerning the integrity of positional information. Even with strategies like AttentionTop that employ a high retention ratio (e.g., 99\%), if the eviction process involves selecting and then compacting non-contiguous token states from a long sequence, the resulting KV cache can lead to severe degradation in generation quality. We attribute this primarily to the disruption of positional encoding integrity. LLMs such as Llama 3 utilize positional encoding schemes like RoPE (Rotary Positional Embeddings) \cite{su2021roformer}, which are learned based on continuous sequences of text. These encodings provide vital signals to the attention mechanism regarding token order, relative distances, and structural relationships within the context. When tokens are removed from various parts of a sequence and the remaining KV states—still carrying their original positional encodings—are effectively concatenated or re-indexed in a compacted cache, the model is presented with a sequence that is, from a positional standpoint, "scrambled." Adjacent states in this modified cache may possess vastly different original positional indices, creating conflicting and misleading signals for the attention heads. This can disrupt the model's ability to track long-range dependencies, understand the narrative or logical flow of the conversation, and ultimately lead to a breakdown in coherence, even if the individually retained tokens had high local attention scores.

The compelling success of the \textbf{SlidingWindowGist} strategy (configured to keep only the first 2000 tokens and discard all recent history) in producing coherent output—precisely when the Baseline was failing on an over-limit context and when AttentionTop (99\% retention) also failed on a near-limit but potentially positionally compromised context—vividly highlights this principle. By providing the model with a shorter but \textit{contiguous and positionally sound} block of initial context, its ability to operate effectively was restored. This suggests that preserving longer, continuous blocks of positional information, even at the significant cost of discarding more total tokens or entire sections of context (such as the middle or most recent history, as in this specific Gist-only configuration), can be substantially more beneficial for maintaining quality than attempting to retain a higher percentage of scattered "important" tokens from a context that is already too long or becomes positionally compromised through the eviction process itself.

Future research must therefore focus on developing eviction strategies that are explicitly aware of and designed to minimize positional disruption. This might involve prioritizing the retention of contiguous blocks of tokens, exploring methods to "heal" or inform the model about the positional gaps created by eviction, or even dynamically adjusting retention policies based on the model's proximity to its architectural limits and the nature of the task. Understanding and navigating the complex trade-off between the sheer amount of context, the continuity and positional integrity of that context, and the inherent architectural limitations of the model is crucial for extending coherent and reliable LLM generation over truly long and complex interactions.

\section{Conclusion and Future Work}
\label{sec:conclusion}
Our stateful benchmarking of KV cache management strategies for `meta-llama/Meta-Llama-3-8b-instruct` reveals that managing context in LLMs is more nuanced than simply pruning tokens to save memory. While the KV cache grows linearly and can become a memory bottleneck, the model's architectural context window (e.g., 8192 tokens) presents an even more critical threshold for maintaining generation quality. Exceeding this limit leads to severe performance degradation.

Furthermore, we find that eviction strategies, even those with high retention rates like 99\% "AttentionTop", can paradoxically harm generation quality if they are applied to already very long contexts and disrupt the integrity of positional encodings by creating non-contiguous effective sequences in the KV cache. The model's reliance on learned positional relationships means that a shorter, positionally coherent context can be superior to a longer but positionally scrambled or architecturally oversized one. This was demonstrated by a simple `SlidingWindowGist` strategy (keeping only the first 2000 tokens) outperforming a Baseline that had exceeded the model's context limit.

These findings emphasize the need for cache eviction strategies that are not only "smart" about content salience but also "structurally aware," prioritizing the preservation of contiguous blocks of context or otherwise minimizing positional disruption. Future work should explore:
\begin{itemize}
    \item Strategies that explicitly balance gist, recency, and (positionally-aware) salience from the middle context.
    \item Methods to quantify and minimize positional disruption during eviction.
    \item The impact of fine-tuning models on post-eviction cache states to improve their robustness to context gaps.
    \item Adaptive thresholds for eviction that consider both cache size and the model's proximity to its effective context coherence limit.
\end{itemize}
Ultimately, enabling truly long, coherent conversations will require a deeper integration of context management techniques with the model's architectural understanding of sequence and position.

\bibliography{bibliography}
\bibliographystyle{acm}
\end{document}